\documentclass{article}

\usepackage{styles/arxiv}

\usepackage[utf8]{inputenc}
\usepackage[T1]{fontenc}
\usepackage{amsmath}
\usepackage{amsfonts}
\usepackage{url}
\usepackage{booktabs}
\usepackage{nicefrac}
\usepackage{microtype}
\usepackage{graphicx}
\usepackage{caption}
\usepackage[numbers, square, sort&compress]{natbib}
\usepackage[table]{xcolor}
\usepackage{tikz}
\usetikzlibrary{positioning,arrows.meta}
\usepackage{tabularx}

\usepackage{hyperref}
\usepackage{doi}

\definecolor{lexiblue}{RGB}{47,105,155}
\definecolor{lexigreen}{RGB}{54,124,94}
\definecolor{lexibluefill}{RGB}{235,243,250}
\definecolor{lexigreenfill}{RGB}{235,246,239}

\title{From Data Room to Report: End-to-End AI Support for M\&A Legal Due Diligence}
\date{}

\usepackage{authblk}

\author[1,2]{Johannes R. Hübers}
\author[1]{Daniel Uedelhoven}
\author[1,2]{Manuela Bergau}
\author[3]{Markus Beyer}
\author[3]{Lars Weber}
\author[3]{Tobias Riethmüller}
\author[3]{Arkadiusz Rzepka}
\author[3]{Tobias Schuh}
\author[3]{Maximilian Glaß}
\author[3]{Omer Faruk Aydin}
\author[3]{Jens Henneberg}
\author[1]{Thiago Bell Felix de Oliveira}
\author[1,2]{Lars Hillebrand}
\author[1,2,4]{Rafet Sifa\thanks{Corresponding author: \texttt{rafet.sifa@iais.fraunhofer.de}, ORCID: 0009-0004-6680-8210.}}

\affil[1]{Fraunhofer IAIS, Sankt Augustin, Germany}
\affil[2]{Lamarr Institute for Machine Learning and Artificial Intelligence, Bonn, Germany}
\affil[3]{G\"ORG Partnerschaft von Rechtsanw\"alten mbB, Cologne, Germany}
\affil[4]{University of Bonn, Bonn, Germany}

\renewcommand{\headeright}{}
\renewcommand{\undertitle}{Accepted at the AI in Legal Industry Workshop, IEEE BigData 2026}
\renewcommand{\shorttitle}{From Data Room to Report: End-to-End AI Support for M\&A Legal Due Diligence}

\hypersetup{
pdftitle={From Data Room to Report: End-to-End AI Support for M\&A Legal Due Diligence},
pdfsubject={cs.CL, cs.AI},
pdfauthor={Johannes R. Hübers, Daniel Uedelhoven, Manuela Bergau, Markus Beyer, Lars Weber, Tobias Riethmüller, Arkadiusz Rzepka, Tobias Schuh, Maximilian Glaß, Omer Faruk Aydin, Jens Henneberg, Thiago Bell Felix de Oliveira, Lars Hillebrand, Rafet Sifa},
pdfkeywords={legal AI, M\&A due diligence, large language models, document classification, information extraction},
}

\begin{document}
\maketitle
\renewcommand{\thefootnote}{}\footnotetext{\copyright~2026 IEEE. Personal use of this material is permitted. Permission from IEEE must be obtained for all other uses, in any current or future media, including reprinting/republishing this material for advertising or promotional purposes, creating new collective works, for resale or redistribution to servers or lists, or reuse of any copyrighted component of this work in other works.}\renewcommand{\thefootnote}{\arabic{footnote}}
\suppressfloats[t]
\begin{abstract}
    Legal due diligence in mergers and acquisitions requires lawyers to organize heterogeneous virtual data room documents, extract transaction-relevant information and consolidate reviewed findings into a due diligence report. We present LEXI (Legal Extraction and Intelligence), a work-in-progress system that demonstrates how large language models can support this process from document ingestion to report preparation. LEXI classifies documents under a configurable due diligence taxonomy, assigns relevant target entities, extracts predefined information with supporting passages and page-level provenance and prepares report content from reviewed findings. Structured outputs connect these stages and allow model-generated results to be passed through the workflow in a controlled form. Lawyers review and correct classifications and extractions and retain responsibility for legal assessments, recommendations, risk ratings and final report wording. Initial evaluations of the classification and extraction components provide preliminary evidence that the approach is feasible. LEXI illustrates how AI can be integrated across the legal due diligence workflow while preserving lawyer control over substantive legal judgment.
\end{abstract}


\section{Introduction and Related Work}
Mergers and acquisitions due diligence is one of the most document-intensive tasks in corporate law. A buyer's legal team receives a virtual data room (VDR) containing shareholder agreements, financial statements, regulatory filings, employment contracts and many other documents. The organization and completeness of these materials vary between transactions, yet review must proceed under tight time and cost constraints. Since seller-provided folder structures do not reliably encode legal relevance, lawyers often need to inspect a document merely to determine its type, responsible workstream and target entity. Before substantive review, they must therefore organize the documents into due diligence areas and route them to the responsible specialists. They can then identify recurring facts, assess their relevance and prepare the findings for the due diligence report.

Developed with a corporate law firm, LEXI supports this workflow by classifying incoming VDR documents under a taxonomy of M\&A due diligence areas, extracting structured answers to predefined information needs and preparing report content from lawyer-reviewed findings. The system proposes document structure and extracted information, while lawyers retain responsibility for legal assessment. Reviewers can confirm, edit or reject the proposed classifications and extractions, add recommendations, assign risk ratings and revise the report draft.

M\&A due diligence has received limited NLP attention despite its document intensity. CUAD~\cite{hendrycks2021cuad}, ContractNLI~\cite{koreeda2021contractnli} and MAUD~\cite{wang-etal-2023-maud} study contract review through clause extraction, document-level entailment and merger-agreement understanding. Their scope is limited to individual English contracts and does not extend to heterogeneous data rooms. Results on English contracts also need not transfer to German-language documents; LLM performance on legal named entity recognition, for instance, varies considerably across datasets from different languages and legal systems~\cite{deusser2024comparative}. Roegiest et al.~\cite{roegiest2018dataset} formulate due-diligence passage retrieval as information retrieval, Jang and Stikkel~\cite{jang2024leveraging} examine LLM support for due diligence, Berger et al.~\cite{berger2023towards} study regulatory-compliance verification and Yang et al.~\cite{10826089} examine generative AI in legal document review. In financial auditing, recommender systems match text passages of financial reports to the legal requirements of accounting standards~\cite{sifa2019towards,hillebrand2023improving}. A zero-shot LLM stage that filters the retrieved passages improves this matching in industrial settings where annotated data for regular fine-tuning are scarce~\cite{hillebrand2023improving}. LEXI carries this requirement-driven review pattern over to heterogeneous data rooms and combines VDR structuring, evidence-backed information extraction, lawyer review and report preparation in one workflow. Its use of zero-shot LLMs with structured outputs relates to recent work on LLM-based text classification~\cite{kostina2025large}, legal-contract classification~\cite{singh2025surveyclassificationtasksapproaches} and generative information extraction~\cite{xu2024large}.

\section{LEXI Workflow}
LEXI processes a VDR in three connected stages. First, it organizes documents under a shared due diligence taxonomy and associates them with the relevant target entities. Second, it extracts the information that specialists need and presents each answer with its source evidence. Third, it organizes the reviewed findings as report content. The stages are linked through reviewed outputs. A confirmed classification determines the applicable extraction catalogue, and only reviewed extractions form the basis of report drafting.

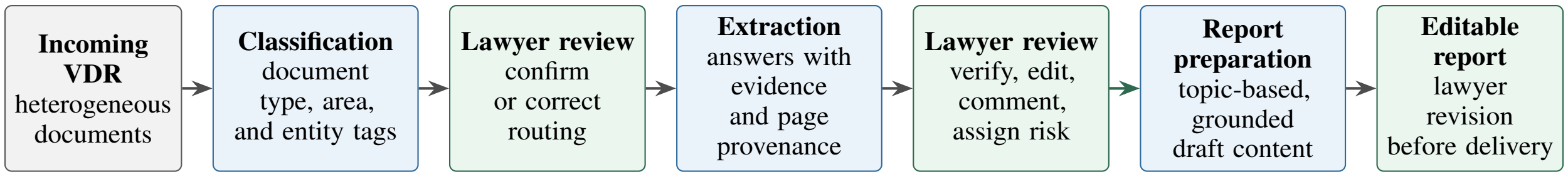
\begin{figure}[!t]
\centering
\begin{tikzpicture}[
    font=\fontsize{8}{9}\selectfont,
    input/.style={draw=black!65, fill=black!5, rounded corners=2pt, align=center, text width=1.65cm, minimum height=1.75cm, inner sep=3pt},
    service/.style={draw=lexiblue, fill=lexibluefill, rounded corners=2pt, align=center, text width=1.95cm, minimum height=1.75cm, inner sep=3pt},
    review/.style={draw=lexigreen!85!black, fill=lexigreenfill, rounded corners=2pt, align=center, text width=1.85cm, minimum height=1.75cm, inner sep=3pt},
    output/.style={draw=lexigreen!85!black, fill=lexigreenfill, rounded corners=2pt, align=center, text width=1.75cm, minimum height=1.75cm, inner sep=3pt},
    flow/.style={-{Stealth[length=2.2mm]}, thick, black!70},
    reviewedflow/.style={-{Stealth[length=2.2mm]}, thick, lexigreen!85!black}
]
\node[input] (vdr) {\textbf{Incoming VDR}\\heterogeneous\\documents};
\node[service, right=3.2mm of vdr] (classify) {\textbf{Classification}\\document type, area,\\and entity tags};
\node[review, right=3.2mm of classify] (route) {\textbf{Lawyer review}\\confirm or correct\\routing};
\node[service, right=3.2mm of route] (extract) {\textbf{Extraction}\\answers with evidence\\and page provenance};
\node[review, right=3.2mm of extract] (review) {\textbf{Lawyer review}\\verify, edit, comment,\\assign risk};
\node[service, right=3.2mm of review] (report) {\textbf{Report preparation}\\topic-based, grounded\\draft content};
\node[output, right=3.2mm of report] (draft) {\textbf{Editable report}\\lawyer revision\\before delivery};

\draw[flow] (vdr) -- (classify);
\draw[flow] (classify) -- (route);
\draw[flow] (route) -- (extract);
\draw[flow] (extract) -- (review);
\draw[reviewedflow] (review) -- (report);
\draw[flow] (report) -- (draft);

\end{tikzpicture}
\caption{LEXI supports the M\&A due diligence workflow from an incoming VDR to report preparation. Automated stages provide structured suggestions, while lawyers control routing, factual review, legal assessment and final report editing.}
\label{fig:lexi-workflow}
\end{figure}

Figure~\ref{fig:lexi-workflow} summarizes the lawyer-facing workflow. LEXI provides structured suggestions and source-grounded information at each automated stage; lawyers control the two review gates and the report's final content.

\subsection{Document Ingestion and Preprocessing}
Documents uploaded to the VDR are processed with OCR and layout
analysis. The resulting document representation retains textual
content together with page and positional information for paragraphs,
tables and other relevant regions. The same representation serves as
model input and supports the review interface. By linking an extracted
statement to a passage on the source page, it allows a lawyer to verify
the proposed answer in context. Retaining these visual cues can also
support the structural analysis of long legal contracts~\cite{hegel2021lawlargedocumentsunderstanding}.

\subsection{Document Classification}
The classification service assigns one or more candidate document-type labels to each VDR document. The taxonomy used in this study reflects German M\&A practice and contains 72 document types grouped into 14 due diligence areas, plus a \texttt{NO\_ACCESS} label for unreadable or DRM-restricted files (Table~\ref{tab:dd-areas}). Each type belongs to one area, and its human-written description records a practitioner-defined legal category. Because the taxonomy is configured when a project or transaction is created, the legal team can adjust it for a transaction or adapt it to other document-review processes. Confirmed classifications support an area-based VDR structure that routes each document to the responsible specialist.

The model receives the taxonomy, including the human-written type descriptions, together with the parsed document text. It returns an ordered list of valid labels under structured output constraints. Classification and extraction operate in a zero-shot setting; the prompts contain no labelled examples. In regulated domains, annotated documents are scarce and privacy-restricted, so that recent work trained a real-estate document classifier entirely on synthetic data~\cite{deusser2025leveraging}. Zero-shot classification needs no training data for any of the 72 types and lets the legal team change the taxonomy without retraining. A single clear label can be applied directly, whereas the review interface presents ranked alternatives when several labels are plausible. The lawyer can accept, edit or reject these alternatives, concentrating manual review on uncertain routing decisions.

\begin{figure}[!b]
\centering
\includegraphics[width=0.8\linewidth]{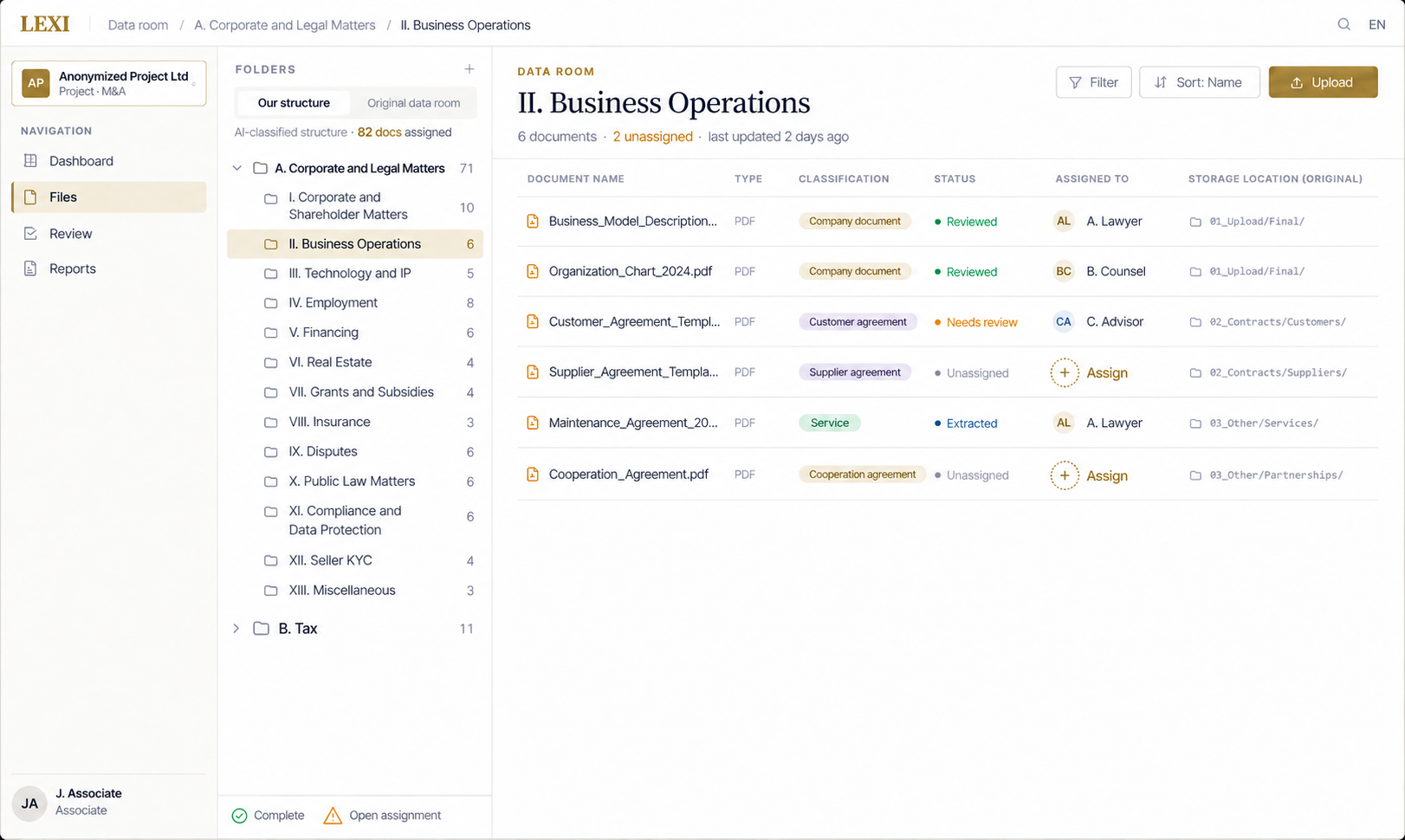}
\caption{Schematic wireframe of the intended document classification interface. The
illustration shows the due diligence structure alongside predicted document
types, review status and lawyer assignments; the production LEXI interface is
still under iteration.}
\label{fig:classification-view}
\end{figure}

The same service also performs \textit{entity tagging}. A target in an M\&A transaction commonly comprises several legal entities that are reviewed in partly separate workstreams. Given the transaction's list of candidate entities, the tagger assigns each document the subset to which it is relevant; a document may concern several entities, one entity or none. The user interface can then filter the VDR by these entity tags, allowing lawyers to focus on the documents associated with a particular target entity while retaining documents relevant across the group.
Figure~\ref{fig:classification-view} illustrates the resulting area-based document view.

\subsection{Information Extraction and Lawyer Review}
\begin{figure}[!t]
\centering
\includegraphics[width=\linewidth]{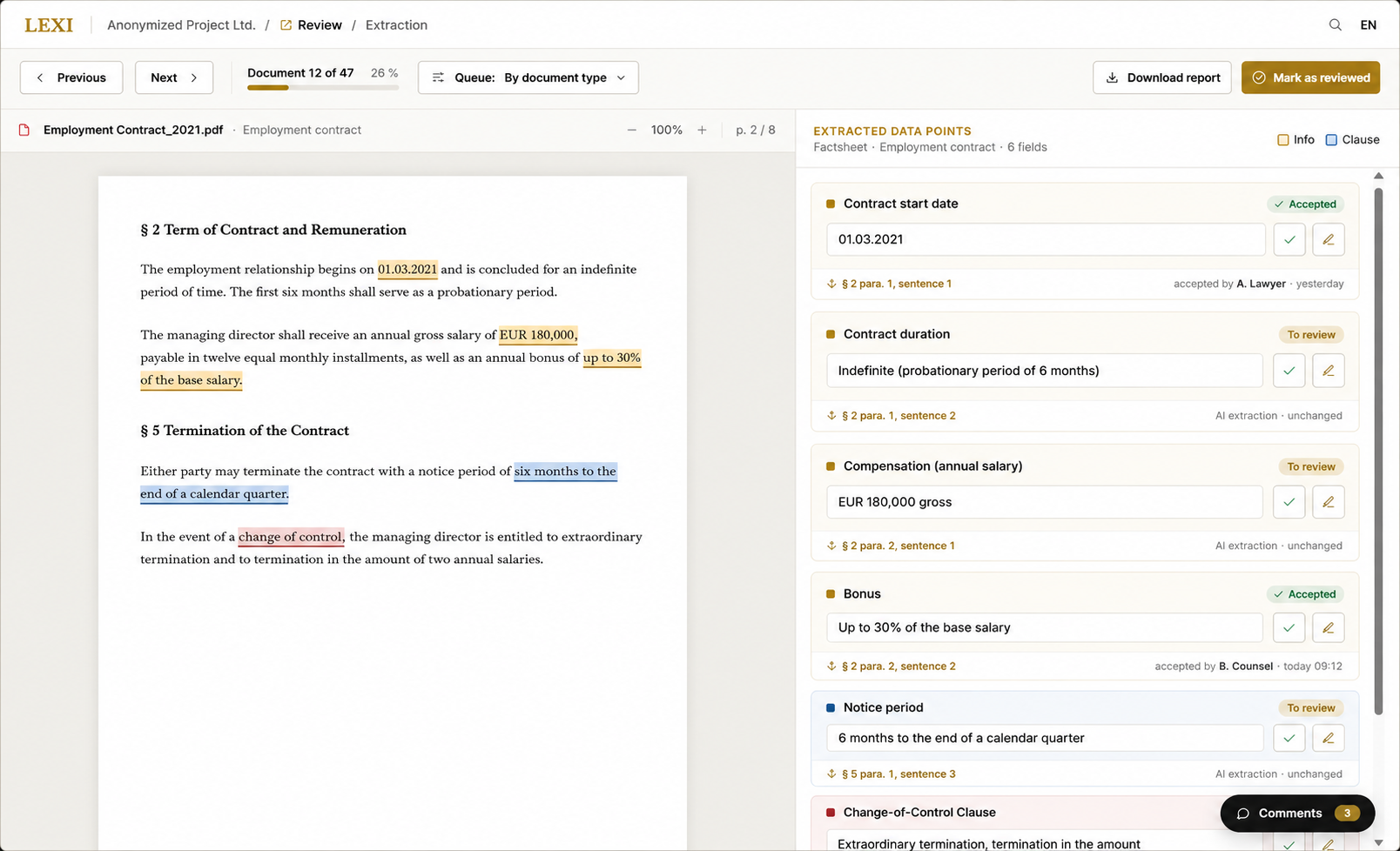}
\caption{Schematic wireframe of the intended extraction review interface. The
illustration shows source-document passages alongside extracted data
points, provenance and lawyer review controls; the production LEXI
interface is still under iteration.}
\label{fig:extraction-view}
\end{figure}

The extraction service answers predefined information needs, called
\emph{extractables}. Each extractable consists of a name and a description of
what to find (for example \emph{Governing Law}). A catalogue of
extractables is associated with each document type, meaning that a confirmed
classification determines which questions are sent to the
extractor with the document text. This avoids applying a single,
generic checklist to every VDR document and lets the legal team adapt
the catalogue to the transaction. Figure~\ref{fig:extraction-view} shows the corresponding review interface.

For every document--extractable pair, the service returns an answer,
an explanation of how the answer was obtained and one or more
verbatim source snippets with page numbers. The answer is labelled
\texttt{exact} when it is copied from the document,
\texttt{inferred} when it requires a deduction or reformatting and
\texttt{not\_found} when the information is absent. The model is
explicitly instructed to return \texttt{not\_found} instead of
guessing. Structured output constraints enforce this response format
and make the evidence, answer and derivation type available as
separate reviewable fields.

Long documents are divided into overlapping chunks of multiple pages
which are then processed through separate parallel LLM calls, thus preserving context
around page boundaries while mitigating LLM context rot. Before returning one answer for each
extractable, another LLM call then reconciles candidates across the windows to
combine complementary evidence and resolve conflicts between parts of
a document.

The source snippets are verified against the parsed document after
generation. A verified passage is mapped to the corresponding lines on
the source page, providing the coordinates used for highlighting in the
review interface. If a snippet cannot be located in the document, it
does not receive a source-page highlight. The interface therefore
distinguishes evidence that can be traced back to the document from an
unverified model-generated citation. This distinction is needed because
answer quality alone does not establish grounding: in regulatory question
answering, a closed-book model that received no source passages scored
nearly as high as a full retrieval-augmented pipeline on an automatic
answer-quality metric, while citing nothing and misstating
obligations~\cite{deusser2026automated}.

The extraction view presents an answer alongside the supporting
passages, including paragraph-level highlighting in the source
document. Lawyers can verify the claim, amend the proposed value, add
a note or recommendation and mark the item as complete or pending.
Once this review is complete, the record retains the corrected value,
the lawyer's comment and the assessment used in report preparation.
Until then, the model output remains a proposed finding and is not
treated as a reportable fact.

\begin{table}[!t]
    \centering
    \caption{M\&A due diligence areas in the document taxonomy.}
    \label{tab:dd-areas}
    \footnotesize
    \begin{tabular}{ll}
        \toprule
        \# & Area \\
        \midrule
        I & Corporate Law and Shareholder Relations \\
        II & Operative Business \\
        III & IT and IP \\
        IV & Employment Law \\
        V & Financing \\
        VI & Real Estate \\
        VII & Grants and Subsidies \\
        VIII & Insurance \\
        IX & Legal Disputes \\
        X & Public Law Relations \\
        XI & Compliance and Data Protection \\
        XII & Seller KYC \\
        XIII & Miscellaneous \\
        XIV & Taxes \\
        \bottomrule
    \end{tabular}
\end{table}

\subsection{Report Preparation}

Due diligence culminates in a report that communicates material
findings, recommendations and risks to the client. Lawyers must turn
many reviewed document-level observations into a consistent report
structure. LEXI's report service uses the reviewed extraction record
for this preparation step. Before generation,
lawyers associate relevant findings with a target entity, add their own
comments or recommendations and assign a risk rating.

Figure~\ref{fig:report-example} shows an illustrative report excerpt
covering two due diligence sections and two target entities. Its topics,
columns and table granularity are chosen by the legal team, and the
service prepares the entries from reviewed findings, comments and risk
ratings.

\begin{figure}[!t]
\centering
\begin{minipage}{\linewidth}
\small
{\scriptsize\color{black!55}\textit{\dots Introduction, subject matter and sources of information \dots}}\\[1mm]
\renewcommand{\arraystretch}{1.10}
\setlength{\tabcolsep}{4pt}
\footnotesize
\fcolorbox{black!50}{black!4}{\parbox{\dimexpr\linewidth-2\fboxsep-2\fboxrule\relax}{\textbf{I. Corporate Law and Shareholder Relations}}}\\[-1mm]
\fcolorbox{black!50}{black!4}{\parbox{\dimexpr\linewidth-2\fboxsep-2\fboxrule\relax}{\textbf{1. Target entity A}}}\\[-1mm]
\begin{tabularx}{\linewidth}{|p{0.55cm}|p{2.1cm}|X|X|p{1.05cm}|}
\hline
\rowcolor{lexiblue}\centering\color{white}\textbf{No.} & \centering\color{white}\textbf{Topic} & \centering\color{white}\textbf{Issue / Implication} & \centering\color{white}\textbf{Notes / Recommendations} & \centering\color{white}\textbf{Risk}\tabularnewline
\hline
1. & \textbf{General} & Registered office and managing directors are recorded. & Confirm against the current register extract. & \centering Low\tabularnewline
\hline
2. & \textbf{Shareholding} & Documents indicate a change in the ownership structure. & Verify transfer documentation and any required consent. & \centering High\tabularnewline
\hline
\multicolumn{5}{|c|}{\scriptsize\color{black!55}\textit{Additional topics continue}}\\
\hline
\end{tabularx}\\[1mm]
\fcolorbox{black!50}{black!4}{\parbox{\dimexpr\linewidth-2\fboxsep-2\fboxrule\relax}{\textbf{2. Target entity B}}}\\[-1mm]
\begin{tabularx}{\linewidth}{|p{0.55cm}|p{2.1cm}|X|X|p{1.05cm}|}
\hline
\rowcolor{lexiblue}\centering\color{white}\textbf{No.} & \centering\color{white}\textbf{Topic} & \centering\color{white}\textbf{Issue / Implication} & \centering\color{white}\textbf{Notes / Recommendations} & \centering\color{white}\textbf{Risk}\tabularnewline
\hline
1. & \textbf{General} & Basic corporate information is available in the reviewed records. & No further action. & \centering Low\tabularnewline
\hline
2. & \textbf{Bodies and powers} & Representation rules and authority requirements are recorded. & Confirm current authority arrangements. & \centering Medium\tabularnewline
\hline
\multicolumn{5}{|c|}{\scriptsize\color{black!55}\textit{Additional topics continue}}\\
\hline
\end{tabularx}\\[1mm]
\fcolorbox{black!50}{black!4}{\parbox{\dimexpr\linewidth-2\fboxsep-2\fboxrule\relax}{\textbf{II. Operative Business}}}\\[-1mm]
\begin{tabularx}{\linewidth}{|p{0.55cm}|p{2.1cm}|X|X|p{1.05cm}|}
\hline
\rowcolor{lexiblue}\centering\color{white}\textbf{No.} & \centering\color{white}\textbf{Topic} & \centering\color{white}\textbf{Issue / Implication} & \centering\color{white}\textbf{Notes / Recommendations} & \centering\color{white}\textbf{Risk}\tabularnewline
\hline
1. & \textbf{Business activity} & Operating information is available across the target group. & Check material customer and supplier dependencies. & \centering Medium\tabularnewline
\hline
2. & \textbf{Material contracts} & Key agreements across the target group require further review. & Confirm assignment and termination provisions. & \centering High\tabularnewline
\hline
\multicolumn{5}{|c|}{\scriptsize\color{black!55}\textit{Additional topics continue}}\\
\hline
\end{tabularx}\\[1mm]
\fcolorbox{black!35}{black!3}{\parbox{\dimexpr\linewidth-2\fboxsep-2\fboxrule\relax}{\centering\scriptsize\color{black!55}\textit{Additional due diligence sections and report tables continue}}}\\[1mm]
{\scriptsize\color{black!55}\textit{\dots Taxes, Scope and conditions and closing report text \dots}}\\[1mm]
{\scriptsize\color{lexigreen!85!black}Illustrative structure only; the system uses no unreviewed model output as a factual basis.}
\end{minipage}
\caption{Illustrative report excerpt. The Corporate Law and Shareholder Relations section is split into entity-specific tables, while Operative Business is shown as one section-wide table. Topics and columns are defined by the legal team; the resulting entries are grounded in reviewed findings and remain editable before delivery. \textbf{Sections:} Corporate Law and Shareholder Relations; Operative Business. Table granularity is chosen per section. Some sections are split by entity, while others cover the target group. The content is anonymized and illustrative.}
\label{fig:report-example}
\end{figure}

\begin{table}[!t]
    \centering
    \caption{Classifier ranking performance with thinking disabled and
    enabled, evaluated against the primary ground-truth label. Each
    entry is the mean of two runs. Hit@$k$ = primary label in the
    top-$k$ predictions; MRR = mean reciprocal rank; labels = mean
    number of labels predicted per document.}
    \label{tab:classifier_eval_A_B}
    \footnotesize
    \begin{tabular}{llccccc@{\hspace{6pt}}ccccc}
        \toprule
        & & \multicolumn{5}{c}{Data room 1} & \multicolumn{5}{c}{Data room 2} \\
        \cmidrule(lr){3-7} \cmidrule(lr){8-12}
        Model & Thinking & Hit@1 & Hit@2 & Hit@3 & MRR & Labels & Hit@1 & Hit@2 & Hit@3 & MRR & Labels \\
        \midrule
        GPT-5.4 & off & 84.2\% & 98.7\% & 98.7\% & 0.914 & 1.37 & 90.4\% & 91.8\% & 91.8\% & 0.911 & 1.06 \\
        GPT-5.4 & on  & 98.1\% & 98.7\% & 98.7\% & 0.984 & 1.14 & 95.2\% & 95.9\% & 95.9\% & 0.956 & 1.08 \\
        \addlinespace[1pt]
        mini    & off & 84.2\% & 95.4\% & 96.1\% & 0.900 & 1.82 & 76.1\% & 83.6\% & 83.6\% & 0.798 & 1.28 \\
        mini    & on  & 88.8\% & 94.7\% & 94.7\% & 0.918 & 1.21 & 95.2\% & 95.2\% & 95.2\% & 0.952 & 1.04 \\
        \addlinespace[1pt]
        nano    & off & 66.5\% & 86.2\% & 89.5\% & 0.775 & 1.90 & 72.6\% & 79.5\% & 80.2\% & 0.763 & 1.37 \\
        nano    & on  & 84.2\% & 87.5\% & 87.5\% & 0.859 & 1.25 & 89.8\% & 90.5\% & 90.5\% & 0.901 & 1.07 \\
        \bottomrule
    \end{tabular}
\end{table}

A confirmed classification provides the first routing step for report
preparation. It identifies the taxonomy area and document types from
which reviewed extraction records can be considered for the
corresponding report section. The report sections therefore map to the
same due diligence areas used to organize the VDR; the selected topics
and rows within a section determine which reviewed findings appear.

For each transaction, the legal team configures the report template,
including the due diligence sections, rows, topics, table columns,
drafting style and report language. The same extraction record can
therefore support the report format required for each mandate. The
report service prepares content for the requested topics and entities
from the lawyer-reviewed findings.

When a report request specifies target entities, the service prepares
content separately for each one. Findings associated with an entity
remain within its report table, while findings without an entity
assignment may be used as group-level information. If the request does
not specify entities for a section, the service generates one table
covering that section's reviewed findings. The legal team can thus set
the granularity by section while preserving entity boundaries where
needed.

The drafting prompt uses the reviewed answer, explanation, comment or
recommendation and risk rating as its factual basis. It directs the
model to select only findings relevant to the requested topic and to
preserve the lawyer's assessment. If the reviewed record contains no
relevant finding, the topic remains empty. Lawyer-supplied comments,
recommendations and risk ratings are binding inputs that the model may
not reassess. These constraints reduce the risk that report generation
introduces unsupported facts or changes the legal team's assessment.

The service exports the structured content to an editable Word report
draft, which lawyers review, revise and supplement before delivery.
This stage reduces the work required to organize reviewed findings,
while the legal team retains responsibility for the analysis and final
wording.

\section{Initial Evaluations}
We evaluated document classification on two small data rooms from real German M\&A transactions. Data room~1 contains 76 documents and data room~2 contains 73. Each document's reference annotation is a list of one or more document-type labels ranked by relevance. We use the first-ranked label as the primary ground-truth label for evaluation. Since the 72 document types have very small per-class support, macro-averaged measures would be unstable. We therefore use ranked-retrieval measures that reflect the routing task. We report Hit@$k$, defined as the share of documents whose primary ground-truth label occurs among the top~$k$ labels, together with mean reciprocal rank (MRR) and the mean number of returned labels. Table~\ref{tab:classifier_eval_A_B} gives results for GPT-5.4, GPT-5.4 mini and GPT-5.4 nano with thinking disabled and enabled (on medium effort). Each entry is the mean of two runs. Enabling thinking increases primary-label Hit@1 in all six model--data-room comparisons and reduces the mean number of returned labels in five (i.e., increases model confidence). Across all configurations, fewer labels are returned for data room~2, and the gains beyond the first label are correspondingly smaller.

Extraction evaluation remains preliminary: a human expert compared 41
German-language document--extractable predictions from GPT-5.4 and
GPT-5.4 mini with reference annotations and found that all 41 matched. This comparison
covers the answer returned for an extractable without comparing source snippets.
Of course, the current sample is too small to support a robust accuracy estimate.
A larger evaluation should assess
answer correctness, the usefulness of the supporting evidence and the
frequency and character of lawyer edits.

The report-preparation stage has not yet been subject to a formal
evaluation. Such an evaluation should assess whether reviewed findings
are placed into the appropriate report topics without altering the
lawyer's assessment. It should also measure report completeness,
drafting quality and the editing effort required from lawyers.

\section{Conclusion}
LEXI supports the M\&A due diligence workflow from VDR structuring through specialist review to report preparation. Across these stages, LLMs propose document classifications and evidence-backed extraction results, while lawyers control routing, assessment and inclusion in the report.

Classification has been evaluated on two data rooms, but the extraction results remain limited to 41 reviewed predictions and report preparation has not yet been evaluated formally. A study of the full workflow with practitioners should therefore measure report quality, review and editing time and performance across additional transactions and jurisdictions. Planned work on chain-of-title analysis will connect ownership and transfer documents across the VDR to reconstruct changes in title and identify gaps for lawyer review. This requires evidence linking across documents and resolution of cross-references among agreements, registers and transfer records. Prior work on zero-shot detection of contradictions and inconsistencies within financial reports~\cite{deusser2023uncovering} provides a starting point for extending such checks from a single report to the documents of a data room. Other planned functions include duplicate and version detection and the matching of contractual amendments to their underlying agreements.

\section*{Acknowledgments}

This research has been partially funded by the Federal Ministry of Education and Research of Germany and the state of North-Rhine Westphalia as part of the Lamarr Institute for Machine Learning and Artificial Intelligence.

\bibliographystyle{styles/unsrtnat}
\bibliography{main}

\end{document}